%% file: main.tex
\documentclass[10pt, a4paper]{article}

\usepackage[final]{lrec2026} 

\usepackage{times}
\usepackage{latexsym}
\usepackage{amssymb}
\usepackage{arydshln}

\usepackage[T1]{fontenc}
\usepackage[utf8]{inputenc}
\usepackage{microtype}
\usepackage{inconsolata}
\usepackage{graphicx}

\usepackage[normalem]{ulem}
\robustify\bfseries
\robustify\uline

\usepackage{float}
\usepackage{caption}
\usepackage{subcaption}
\usepackage{natbib}
\usepackage{hyperref}
\usepackage{listings} 
\usepackage{tcolorbox} 
\usepackage{makecell} 
\usepackage{multirow}
\usepackage{tabularx}
\usepackage{diagbox}
\usepackage{csquotes}

\usepackage{siunitx}
\usepackage{enumitem}
\usepackage{amsmath}
\usepackage{booktabs}
\usepackage{cleveref}

\usepackage{tablefootnote} 

\title{Towards Empowering Consumers through Sentence-level Readability Scoring in German ESG Reports}

\name{Benjamin Josef Schüßler$^\diamondsuit$, Jakob Prange$^\spadesuit$} 

\address{$^\diamondsuit$University of Augsburg, Germany \\
$^\spadesuit$German Center for Addiction Research in Childhood and Adolescence (DZSKJ),\\ University Medical Center Hamburg-Eppendorf, Germany \\
         benjamin.schuessler@uni-a.de, j.prange@uke.de}

\abstract{
With the ever-growing urgency of sustainability in the economy and society, and the massive stream of information that comes with it, consumers need reliable access to that information.
To address this need, companies began publishing so called Environmental, Social, and Governance (ESG) reports, both voluntarily and forced by law. To serve the public, these reports must be addressed not only to financial experts but also to non-expert audiences. But are they written clearly enough?
In this work, we extend an existing sentence-level dataset of German ESG reports with crowdsourced readability annotations. We find that, in general, native speakers perceive sentences in ESG reports as easy to read, but also that readability is subjective.
We apply various readability scoring methods and evaluate them regarding their prediction error and correlation with human rankings. 
Our analysis shows that, while LLM prompting has potential for distinguishing clear from hard-to-read sentences, a small finetuned transformer predicts human readability with the lowest error. Averaging predictions of multiple models can slightly improve the performance at the cost of slower inference.\textsuperscript{1}
 \\ \newline  }

\begin{document}

\maketitleabstract

\stepcounter{footnote}
\footnotetext{Code and dataset extension available at:\\\href{https://github.com/schuesslerbenjamin/Sentence-level-Readability-Scoring-in-German-ESG-Reports}{github.com/schuesslerbenjamin/Sentence-level-Readability-Scoring-in-German-ESG-Reports}. Trained models available at:\\\href{https://huggingface.co/schuesslerbenjamin/Sentence-level-Readability-Scoring-in-German-ESG-Reports}{huggingface.co/schuesslerbenjamin/Sentence-level-Readability-Scoring-in-German-ESG-Reports}}

\section{Introduction}

In order to make transparent how corporate economic goals align with, contribute to, or violate sustainability goals, policymakers demand written reporting on environmental, social, and governance topics, in short, ESG reports.\footnote{EU Directive 2022/2464, also known as the Corporate Sustainability Reporting Directive (CSRD):\\\href{https://eur-lex.europa.eu/eli/dir/2022/2464/oj/eng}{eur-lex.europa.eu/eli/dir/2022/2464/oj/eng}}
Next to \textit{greenwashing} \citep[the intentional or negligent misrepresentation of one's sustainability strategy to sound more positive and marketable than it really is,][]{de2020concepts}, another challenge is ensuring the reports' accessibility to their diverse audiences. This is even more important for layperson consumers than for other stakeholder groups such as economic auditors or financial analysts.
The latter know exactly what they are looking for and, in case of unclear language, can consult with legal or public relations experts. This is usually not the case for consumers, who may be on their own and may read exploratorily, to gather information from scratch.
Quoting EU Directive 2024/825, also known as the Empowering Consumers Directive,\footnote{\href{https://eur-lex.europa.eu/eli/dir/2024/825/oj/eng}{eur-lex.europa.eu/eli/dir/2024/825/oj/eng}}
\textit{
    ``[i]n order to contribute to the proper functioning of the internal market, based on a high level of consumer protection and environmental protection, and to make progress in the green transition, it is essential that consumers can make informed purchasing decisions and thus contribute to more sustainable consumption patterns. That implies that traders have a responsibility to provide \textbf{clear, relevant and reliable information}.''
}

\begin{figure}[t]
    \centering
    \includegraphics[width=\columnwidth]{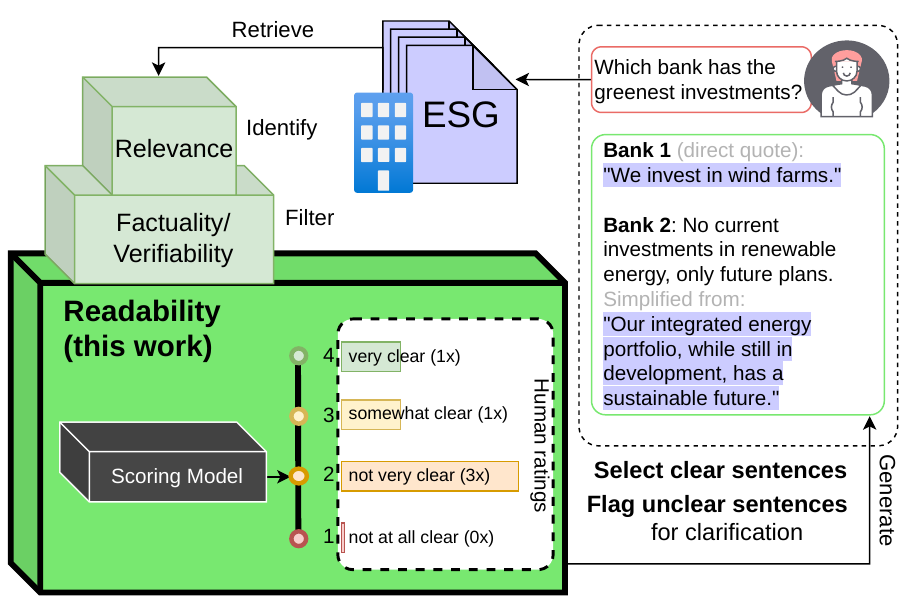}
    \caption{Readability as a foundation of consumer empowerment from ESG reports.}
    \label{fig:teaser_header}
\end{figure}

In this work, we focus on clarity as a fundamental requirement for consumer accessibility, and evaluate automatic readability scorers against the judgments of layperson readers (figure \ref{fig:teaser_header}).

Automatic readability assessment (ARA) is the task of estimating how easy a text is to read and understand. Often, this is measured at the document level using broad sentence and word length statistics to assess generic readability. For example, the Flesch Reading Ease test \cite{flesch_new_1948} is intended to rate English educational books on a scale ranging from school grades to professional scientist difficulty.

Here, on the other hand, we are interested in measuring readability in German ESG reports and in more fine-grained grammatical patterns than length. As our target group, we envision, for example, a young adult deciding on a sustainable bank to open their first account or investment plan with, or a family choosing an electricity provider for their home.
Already overwhelmed with the multitude of companies to choose from and only able to skim very short excerpts of reports from each company,
they rely on a retrieval-augmented generation \citep[RAG; cf.][]{kleinle-etal-2024-omos} or recommender system \citep[cf.][]{hillebrand_etal_2023_sustainai}. We consider a previously unaddressed requirement for such a system, namely to maximize the clarity of the presented content: systems should (a) prefer easily readable sentences for direct extraction and (b) if a sentence is highly relevant but difficult to read, it should be simplified.
Rather than trying to replicate coarse document-level scores like Flesch, we thus propose to model readability at the sentence level.

We aim to answer two questions:
\textbf{RQ1:} How readable are German sustainability reports?
And \textbf{RQ2:} How to model sentence-level readability?
Concretely, we contribute:
\begin{itemize}
    \item an in-depth data analysis of German ESG reports through a crowd-sourcing annotation study, finding largely clearly written sentences but also subjective variation;
    \item a comparison of different model types, including generative Large Language Models (LLMs), regressions, and a custom feature-based classifier, finding lower prediction error in small finetuned models and higher ranking correlation, albeit on a shifted scoring scale, in one of the LLMs;
    \item an ablation of syntactic features, highlighting their relative importance in sentence-level readability prediction;
    \item and a discussion of sentence-level readability in the context of other factors of consumer empowerment through ESG reporting.
\end{itemize}

\section{Related Work}

To find similar research, we systematically queried the typical research databases (see Appendix \ref{appendix:literature_search}).

\paragraph{Readability of German texts.}

While most research on readability has focused on English texts \cite{collins-thompson_computational_2014}, some approaches have also been adapted to the German language. \citet{amstad_wie_1978}, for example, adapts the Flesch Reading Ease formula by \citet{flesch_new_1948} by changing the factor for the word length to consider that German words tend to be longer. More recent research includes creating more sophisticated readability formulae (e.g, \enquote{Hohenheimer Komplexitätsindex für Politikersprache} (HKPS, German for Hohenheim Complexity Index for Political Language) by \citet{kercher_verstehen_2013}), improving the readability for people with learning difficulties (e.g., \citet{jablotschkin-etal-2024-de}), or analyzing how difficult language learners perceive the readability of texts (e.g., \citet{weiss-meurers-2022-assessing}). 

Furthermore, the GermEval 2022 shared task on text complexity assessment of German texts by \citet{salar-mohtaj-babak-naderi-2022-overview} is based on sentences from articles in the areas of society, science, and history of the German Wikipedia. It motivated a wide range of approaches, the best of which was an ensemble of GBERT and GPT-2 submitted by \citet{blaneck-etal-2022-automatic} and achieved a 0.195 MSE (0.442 RMSE) on a 7-point rating scale.

Our study, instead, focuses on the readability as perceived by \textit{native speakers} who are laypersons in the \textit{ESG domain}.

\paragraph{Readability of ESG reports.}

\citet{smeuninx_measuring_2020} compare the performance of traditional readability formulae with a few modern NLP methods when predicting the readability of English ESG reports. They find that the former lack in performance, especially when the syntax varies. In general, \citet{smeuninx_measuring_2020} identify that ESG reports can be difficult to read, in some instances even more complex than financial reporting.

Among other linguistic aspects, \citet{huang_textual_2024} analyze the readability of Chinese ESG reports and their impact on the ESG scores over time.
\citet{bonn_does_2024} investigate how report readability, among other parameters, correlates with the overall sustainability of German companies (``ESG-Score'' assigned by auditors), but they do not \textit{predict} readability and the reports they analyze are written in English.

\paragraph{Methodologically} very similar to our work are \citet{vajjala-meurers-2012-improving}, who compare syntactic features against ``traditional features'' like word length and sentence length and achieve 0.023 MSE (0.15 RMSE) on a 5-point rating scale. But they, again, work with English texts in the educational domain rather than German ESG reports.

\pagebreak

\section{Data}
\label{Data}

For our experiments, we use the dataset\footnote{\href{https://github.com/SustainEval/sustaineval2025\_data/}{github.com/SustainEval/sustaineval2025\_data/}}
from the SustainEval GermEval shared task on understanding sustainability reports
\citep{prange-etal-2025-overview}. It consists of short excerpts sampled from the German Sustainability Code (Deutscher Nachhaltigkeitskodex),\footnote{\href{https://www.deutscher-nachhaltigkeitskodex.de/en/}{deutscher-nachhaltigkeitskodex.de}}
where companies can voluntarily publish ESG reports and receive feedback and resources to prepare for legally required and audited CSRD reporting.
Specifically, each datapoint consists of four consecutive sentences in German, of which the last is the target sentence receiving annotation and the preceding ones are provided for context.
Statistics are given at the top of table~\ref{tab:dataset_metrics}.

\nocitelanguageresource{sustaineval-2025-data}

\paragraph{Readability Annotation.}

We extend the ``verifiability'' annotations used for the SustainEval shared task with layperson readability judgments via crowd-sourcing.
Training and evaluation crowd annotators were recruited via Prolific and paid above German minimum wage. The actual annotation was carried out via SoSciSurvey on GDPR-compliant servers in Germany. Development crowd annotation was carried out on a different platform, also according to German minimum wage and GDPR standards. The change in annotator pools likely led to the difference in agreement and score distributions.
In all cases, the only information disclosed by annotators was that they speak German as their primary language. Annotators were identified only by anonymous IDs, which enabled us to exclude annotators from future annotation rounds if they were too fast or always assigned the same category.
While the three context sentences were shown, annotators were asked specifically to rate their understanding of only the target sentence on a forced-choice Likert scale (\textit{How well do you understand the sentence?} 1: not at all, 2: rather not, 3: somewhat, 4: very clearly).

\input{tables/data_table_complete}

\paragraph{Agreement.}
Most sentences were rated by 5 annotators (some by 4 and very few by 6), and most of the time (72.3--87.7\%), a majority of at least 3 annotators assign the same rating (middle part of table \ref{tab:agreement}).
Due to the lack of annotator identities in the data, we were not able to compute chance-corrected agreement metrics such as Cohen's $\kappa$ or Krippendorff's $\alpha$.
To gain a more comprehensive measure of agreement, we introduce \textit{Mode Agreement}, which cleanly handles anonymity and varying numbers of annotations per sentence.
For each sentence, we count how many annotators agree on the most common rating (the mode), and divide by the number of annotations that sentence received (see equation \ref{eq:Mode_Agreement}). If all annotators agree, the Mode Agreement is one. If no annotators agree, the Mode Agreement is zero.

\begin{equation}
\small
\parbox[c]{1.0cm}{Mode Agreement} = \begin{cases} \frac{\text{Mode's Frequency}}{\text{\# Annotations}},& \text{if Mode's Frequency} \geq 2\\ 0, & \text{else.}\end{cases}
\label{eq:Mode_Agreement}
\end{equation}

Weighing the majority agreement in this way, while still in an acceptable range of 60.4--70.3\%, paints a somewhat less optimistic picture than simply counting how often a majority exists (see table~\ref{tab:agreement}).
Based on this, we decided to account for outlier noise in crowd-sourcing by aggregating the annotations using the majority vote instead of the mean over all votes. Only in the case of ties, we take the mean of the tied votes.

\begin{figure*}[t]
    \centering
    \includegraphics[width=\linewidth]{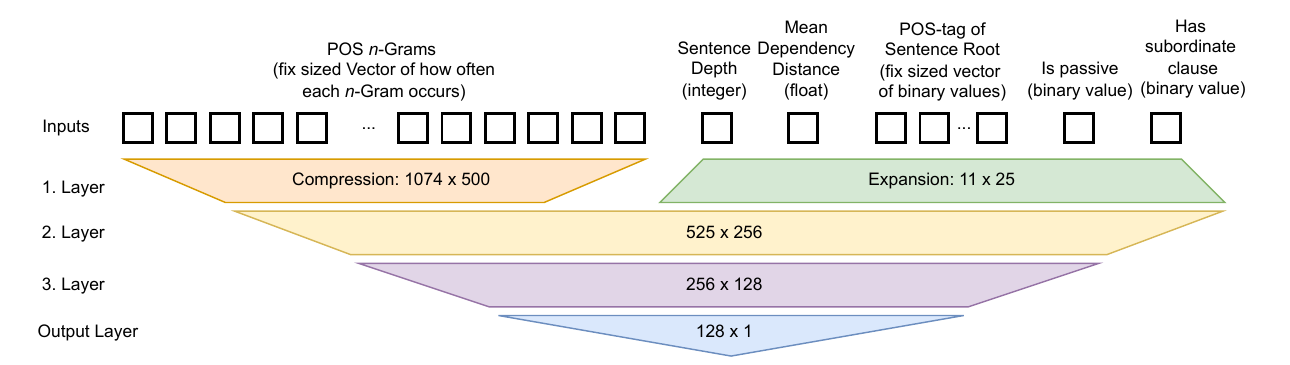}
    \caption{Structure of our syntax-based ARA model.}
    \label{fig:ARA-Syntax-structure}
\end{figure*}

\paragraph{Score distribution.}

The bottom part of table \ref{tab:understandability_annotations} shows that the annotations are skewed towards very easily readable texts.
This is most substantial in the training and evaluation splits, where more than two thirds of all instances were assigned the highest readability score of 4. This can have several reasons: Firstly, the annotators are self-declared native laypeople.
We expect that most laypeople generally perceive texts in their native language as at least somewhat readable. Secondly, crowd-workers might fear that if they rate a text as not understandable, they might not be allowed to answer the other questions and get paid less. To ensure models learn to predict scores across the full scale during training, we randomly oversample the underrepresented rating classes in the training split until they match the most common class. We do not manipulate the distribution in the development and evaluation splits.
Scores are normalized to [0.0; 1.0] for model training, inference, and evaluation.

\section{Whitebox Readability Models}
\label{Automatic Readability Assessment}

In line with our goal to predict sentence-wise readability in ESG reports in a transparent and interpretable manner, we explore a simple but effective whitebox model, where input features can be controlled, parameter sizes are small, and runtimes are fast (section~\ref{sec:features}).
We hypothesize that to capture actual human judgments in a specialized domain, established formulae with linear coefficients preset to a fixed educational setting are not sufficient and test this with a baseline model (section \ref{sec:method-baselines}).
On the other hand, state-of-the-art language models (section \ref{sec:method-blackbox}) may be more powerful than we need and their large blackbox architectures restrict the linguistic insight they may provide.

Given the use case outlined in the introduction, where a system needs to judge how presentable individual sentences are to a layperson user, we focus on the lower-bound setup of providing only the target sentence to the models, without context.
Within-sentence grammatical patterns are complementary to across-sentence semantic and pragmatic aspects of readability like verifiability, coherence, and cohesion.
These aspects likely also have a large impact on readability and are modeled to various extents by pretrained (L)LMs. By examining quantitative and qualitative differences in readability scores assigned by the various models, we can approximate which facets of readability can be determined from syntax alone and which stem from other linguistic properties.

\subsection{Syntactic Features}\label{sec:features}

Inspired by the work of \citet{liu2025automatic} and \citet{smeuninx_measuring_2020}, we design a feed-forward neural network on top of syntactic features extracted from the input sentence (figure \ref{fig:ARA-Syntax-structure}). Most notably, to limit the impact of the Part-of-Speech (POS)-tag $n$-Grams, their vector is compressed in the first layer before being concatenated with the other features. For more implementation details, see appendix \ref{appendix:syntax-implementation}. 

Our model uses the following features:
\paragraph{Part-of-Speech-tag $n$-grams.} The first syntactical feature analyzes the grammatical structure of a sentence, as it can have an impact on the syntactical complexity and, thus, on the readability of a sentence \cite{razon_new_2015}. This is based on the idea that if a sentence structure is observed more often, a reader is more likely to understand it easily \cite{kauchak_measuring_2017}. Using a sliding window, we represent a shallow view of a sentence's syntactic structure as count features of POS-tag bigrams and trigrams.
We filter for punctuation symbols.
Before training, we generate all bigrams and trigrams that appear in the training set and expand them into individual features, representing how often each bigram and trigram appears in the sentence. 
From the training data, we extract 158 unique bigrams and 916 unique trigrams resulting in 1,074 total $n$-gram feature dimensions.

\paragraph{Depth of the dependency tree.} As proposed by \citet{yngve_model_1960}, we calculate the depth of the dependency tree to estimate the hierarchical complexity of a sentence. 
In a dependency tree, every word except for the root has exactly one head that it refers to. The deeper the dependency tree of a sentence, the higher its syntactic complexity, and thus---we hypothesize---the harder it is for the reader to understand the relations of the words within the sentence.

\paragraph{Mean dependency distance.} We also calculate the mean dependency distance as proposed by \citet{haitao_liu_dependency_2008}. The dependency distance is thereby defined as the number of words between a word and its head. Using the mean dependency distance instead of summing up all distances within a sentence prevents longer sentences from getting disproportionally higher scores \cite{haitao_liu_dependency_2008}.

\paragraph{Part-of-Speech-tag of the root.} There is exactly one word in every sentence that has no head in the dependency tree, the root. It can have a major impact on a sentence's readability \citep{dellorletta_readit_2011}. 
We extract all root POS-tags within the training data and expand them into binary variables. We find that verbs or auxiliary verbs are usually the root of sentences in our dataset. Thereby, our approach allows the model to find relations between all possible POS-tags of sentence roots and the readability of a sentence.

\paragraph{Passive voice.} Sentences written in passive voice can also be harder to read. Thus, \citet{smeuninx_measuring_2020} analyze the readability of whole documents and calculate the proportion of sentences that are in passive voice. Since we are only working with single sentences, we create a binary variable indicating whether the text is in passive voice. We consider a sentence to be passive voice if it contains a participle that has a form of \enquote{werden} (the equivalent of passive \textit{to be}) as its head, or if it includes a passivized subject.

\paragraph{Subordination.} Finally, sentences consisting of multiple clauses can be more complicated than sentences with fewer clauses. \citet{smeuninx_measuring_2020} calculate the average number of subclause-introducing elements per sentence to represent the degree of subordination in a document. We adopt their idea to our sentence-level ARA task and create a binary variable that indicates whether there is at least one subordinate conjunction in the sentence.

\subsection{Baselines}\label{sec:method-baselines}

To test whether our selected linguistic features are more informative in our setting than established work on readability suggests, we compare with two baselines. Both baselines are trained on our German ESG-report data to account for domain effects.

\paragraph{Sentence Length.}
We train a simple linear regression model using only the number of words per sentence. This approach was used by \citet{crossley_toward_2007} as a proxy for syntactic complexity. 

\paragraph{Readability Formulae.}
Representing the traditional research on readability, we train an XGBoost model \cite{chen_xgboost_2016} over scores calculated using established readability formulae.\footnote{We also experimented with aggregating the scores using Linear Regression, Ridge Regression, Lasso Regression, and Elastic Net, but XGBoost led to the best results overall.}
We select the following formulae due to their relevance and applicability to German sentences: 
the \textit{Flesch-Reading-Ease} test introduced by \citet{flesch_new_1948} for English texts and adapted to German texts by \citet{amstad_wie_1978};
the \textit{Hohenheim Complexity Index for Political Language} \cite[HKPS,][]{kercher_verstehen_2013}; the proportion of polysyllabic words, based on the idea of the \textit{Gunning Fox} \cite{gunning_technique_1952} and \textit{SMOG indices} \cite{mclaughlin_smog_1969}; the \textit{Vienna Educational Text Formula} \citep{bamberger_lesen-verstehen-lernen-schreiben_1984}; and the Swedish readability index LIX \citep{bjornsson_lasbarhet_1968}.
See appendix \ref{appendix:rf-model} for details on the formulae and their implementation.

\section{Blackbox Readability Models}\label{sec:method-blackbox}

As reference, we also compare the syntactic features model with two types of modern language models: a finetuned classifier on top of a pretrained transformer encoder and instruction-tuned generative LLMs. This is to set a practical upper bound in terms of predictive power. If a whitebox model reaches or surpasses the blackbox models' prediction accuracy, the whitebox model should clearly be preferred. Otherwise, a tradeoff between accuracy, speed, and interpretability needs to be found.

\subsection{XLM-RoBERTa Encoder-Classifier}

For the first language model, we use a transformer encoder model and task-specifically finetune it to the ARA task. This approach follows \citet{tseng_innovative_2019} and can simultaneously consider several linguistic layers of a text, including semantic and syntactic aspects, making it more powerful in principle than the syntax-based model. Since we define ARA as a regression task, we train the model's final layer as a regression head (see appendix \ref{appendix:RoBERTa-implementation} for more details). 

This allows the model to effectively predict readability scores on a scale from zero to one. We compare several BERT-like encoder models on the development set, and select the multilingual \texttt{XLM-RoBERTa-base}\footnote{\href{https://huggingface.co/FacebookAI/xlm-roberta-base}{huggingface.co/FacebookAI/xlm-roberta-base}}
and \texttt{XLM-RoBERTa-large}\footnote{\href{https://huggingface.co/FacebookAI/xlm-roberta-large}{huggingface.co/FacebookAI/xlm-roberta-large}}
models \citep{conneau_unsupervised_2020,liu_roberta_2019} based on their performance. 

\input{tables/results_reordered}

\subsection{Generative LLMs}

For the second language model, we test instruction-tuned LLMs on the ARA task, focusing on instruction-tuned models pretrained on datasets that include German texts. 

We compare the \texttt{Llama 3 8B instruct} model\footnote{\href{https://huggingface.co/meta-Llama/Llama-3.1-8B-Instruct}{huggingface.co/meta-Llama/Llama-3.1-8B-Instruct}}
by \citet{Llama_team_ai__meta_Llama_2024} with the \texttt{Qwen 3 4B Instruct 2507} model\footnote{\href{https://huggingface.co/Qwen/Qwen3-4B-Instruct-2507}{huggingface.co/Qwen/Qwen3-4B-Instruct-2507}}
by \citet{qwen_team_qwen3_2025} and the \texttt{Gemma 3 4B it} model\footnote{\href{https://huggingface.co/google/gemma-3-4b-it}{huggingface.co/google/gemma-3-4b-it}}
by \citet{gemma_team_gemma_2025}. We analyze these LLMs, as they are highly relevant in the current research and have been extensively researched.

We prompt the models with similar instructions as the human annotators. While the dataset itself is in German, we prompt the models in English as previous research found that several LLMs are biased to internally pivot towards English due to imbalanced training data \cite{wendler_Llamas_2024}. We instruct the models to classify the readability of a sentence into four classes, each coded with a number from one through four. Similar to the human annotations, these numbers are then scaled down to the same range from zero to one. We apply one-shot prompting (see appendix \ref{appendix:llm-prompt}), because early experiments on the development data split showed that zero-shot prompts lead to worse performance. For the shots, we randomly sample sentences and their readability score from the training data.

\section{Experimental Results}

All performance metrics can be found in table~\ref{tab:results}, while table~\ref{tab:average_predictions} contains average predicted scores. Details on the experimental setup can be found in appendix \ref{appendix:experimental-setup}.

\paragraph{Metrics.}

We use Mean Squared Error (MSE) as our main metric for analysis and as the loss function during training.
We also report the Mean Absolute Error (MAE), as it is more robust to outliers than the MSE, and more interpretable because it is true-to-scale. For both error metrics, a lower score is considered better, where 0 is the best possible error and 1 is the worst possible.

\input{tables/average_predictions}

Additionally, we report a rank order correlation score to analyze whether a model can correctly identify which sentences are easier to read than others. The sentences are sorted by the predicted scores, and then this order is compared to the sorted list of gold-standard annotations \cite{collins-thompson_computational_2014}.
Rank correlation measures the extent to which the predicted order aligns with human annotations across entire datasets. 
A model that can distinguish between easy and hard-to-read sentences but has a systematic bias to too low or high scores, provides more value than a model that makes incorrect predictions in both directions. MSE does not capture this difference, which is why a rank order correlation score is needed. 
We use the Kendall $\tau$ coefficient, variant b, introduced by \citet{kendall_treatment_1945} as it is often used in the research and accounts for ties.
To calculate the scores, we use the Python implementation by the library \texttt{scipy} by \citet{jones_scipy_2001}\footnote{We also experiment with rounding the predictions to the next .5 before calculating the Kendall $\tau$ score, but as this leads to similar results, we keep the original calculation.}. A Kendall $\tau$ score of $+$1 indicates perfect correlation or correct ranking, relative to the ground truth. A score of 0 indicates no correlation, and $-$1 shows that the rankings are inverted.

\subsection{Individual Models}

\paragraph{Sentence Length.} 
This simple baseline yields mediocre error metrics and the Kendall $\tau$ score is worse than random, indicating that the model fails to distinguish easy from hard-to-read sentences. Further, this shows that sentence length by itself is not a good predictor of readability.

\paragraph{Readability Formulae.}
Aggregating the scores of several traditional readability formulae leads to better results than the simple sentence-length baseline, according to all metrics. This shows that the additional sentence parameters and weightings used in readability formulae allow for a better prediction of the readability than relying on the number of words alone.

\paragraph{Syntactic Features.}
With our proposed feature-based approach, we aim to predict sentence readability by having the model learn to analyze several syntactic patterns. This model is more complex than the two baselines, both in terms of its input features and degrees of freedom of its hidden layers.
This added representational capacity leads to a similar error rate and ranking performance as the formulae.

To investigate the importance of individual features, we conduct an ablation study (table \ref{tab:results_ARA-Syntax-ablation}). A feature is more important to the model if its removal strongly negatively affects the performance, i.e. increases the error rate or lowers the Kendall $\tau$ score. Removing trigrams worsens the performance the most, according to all three metrics. Thus, it is the most influential feature. Passivization (by the error metrics) and the depth of the sentence (by the Kendall $\tau$ score) are, respectively, the second most important features.

Interestingly, removing the bigrams very slightly improves the performance of the model, according to the error metrics. A possible reason is that the information in the bigrams is already part of the trigrams and the model compresses the $n$-gram input vector to a fixed width. However, according to the Kendall $\tau$ scores, ablating any feature leads to worse performance, indicating their necessity to correctly distinguish easy from hard sentences.

\input{tables/ARA-Syntax-ablation}

\paragraph{XLM-RoBERTa.}

We observe a strong improvement over the syntax model in all metrics. This can be argued with the transformer model's higher degrees of freedom to fit to the task. The XLM-RoBERTa model thereby predicts with the smallest errors of all individual models.
We also find, rather surprisingly, that the large model variant performs much worse than the base model variant, only slightly beating the simple sentence length baseline. This may be due to insufficient training conditions that fail to saturate the many parameters of the large model.

\paragraph{Generative LLMs.}
Prompting different LLMs, we find that Llama and Gemma fail to correctly estimate the readability, as indicated by high error metrics and very low Kendall $\tau$ scores. Qwen outperforms the other LLMs in every metric. Thus, we use Qwen as the representative LLM going forward. The highest Kendall $\tau$ score out of all the individual models indicates that the LLM is better at distinguishing easy from hard-to-read sentences than XLM-RoBERTa and the whitebox models. However, as it has not been finetuned on our specific dataset and rating scale, it is much worse than our syntax model, the readability formula model, and XLM-RoBERTa at assigning scores that are numerically close to the human ratings. Specifically, it assigns lower scores on average than most other models and humans (table~\ref{tab:average_predictions}).

\begin{figure}
    \centering
    \includegraphics[width=\linewidth]{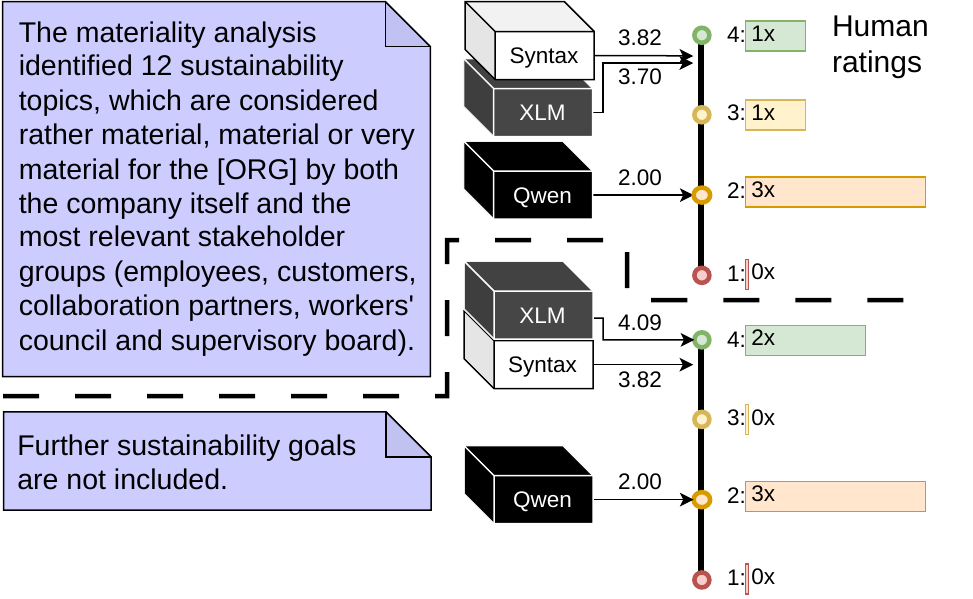}
    \caption{Two examples of different length and superficial complexity (translated from German).}
    \label{fig:examples}
\end{figure}

\paragraph{Error Analysis.}

Consider the examples in figure \ref{fig:examples}.
The top sentence is lengthy and syntactically complex.
There is some variation in human judgments but consensus is clearly ``not very readable''.
This is reflected in all model predictions being less than the top score, though the syntax model and XLM-RoBERTa are still (too) optimistic, while Qwen matches the human vote.
The bottom sentence is short and not complex, but understanding it requires access to the preceding context, which was provided to humans but not to models.
This lead to a bimodal distribution in human judgments, as some annotators likely focused on the low syntactic complexity (high readability) while others emphasized context-dependence (low understanding).
Models diverge similarly, and Qwen again happens to match the majority vote which in this case is only narrowly decided.
Note that Qwen and the Syntax model assign the same score to these two examples.

\subsection{Model Combinations}

To account for different aspects of readability influencing individual models differently, we also experiment with averaging the predictions of the three models.\footnote{Additionally, we tried other aggregation methods, including Linear Regression, Ridge Regression, and XGBoost. However, all models performed similarly, thus we chose mean aggregation for simplicity.}
The combination of the syntactic model and RoBERTa has the lowest MSE (but not MAE)\footnote{MSE is more sensitive than MAE to individual datapoints with large errors. So the difference between the model with the lowest MSE and the model with the lowest MAE lies in deviating less from the ground truth on outliers versus getting the majority of the data closer to it.} out of all experimental settings, but only by a small margin. Combining Qwen's and RoBERTa's predictions slightly outperforms Qwen's individual Kendall $\tau$ score. 
Overall, simple mean aggregation does improve predictions slightly, but not substantially.

\begin{figure}
    \centering
    \includegraphics[width=\linewidth]{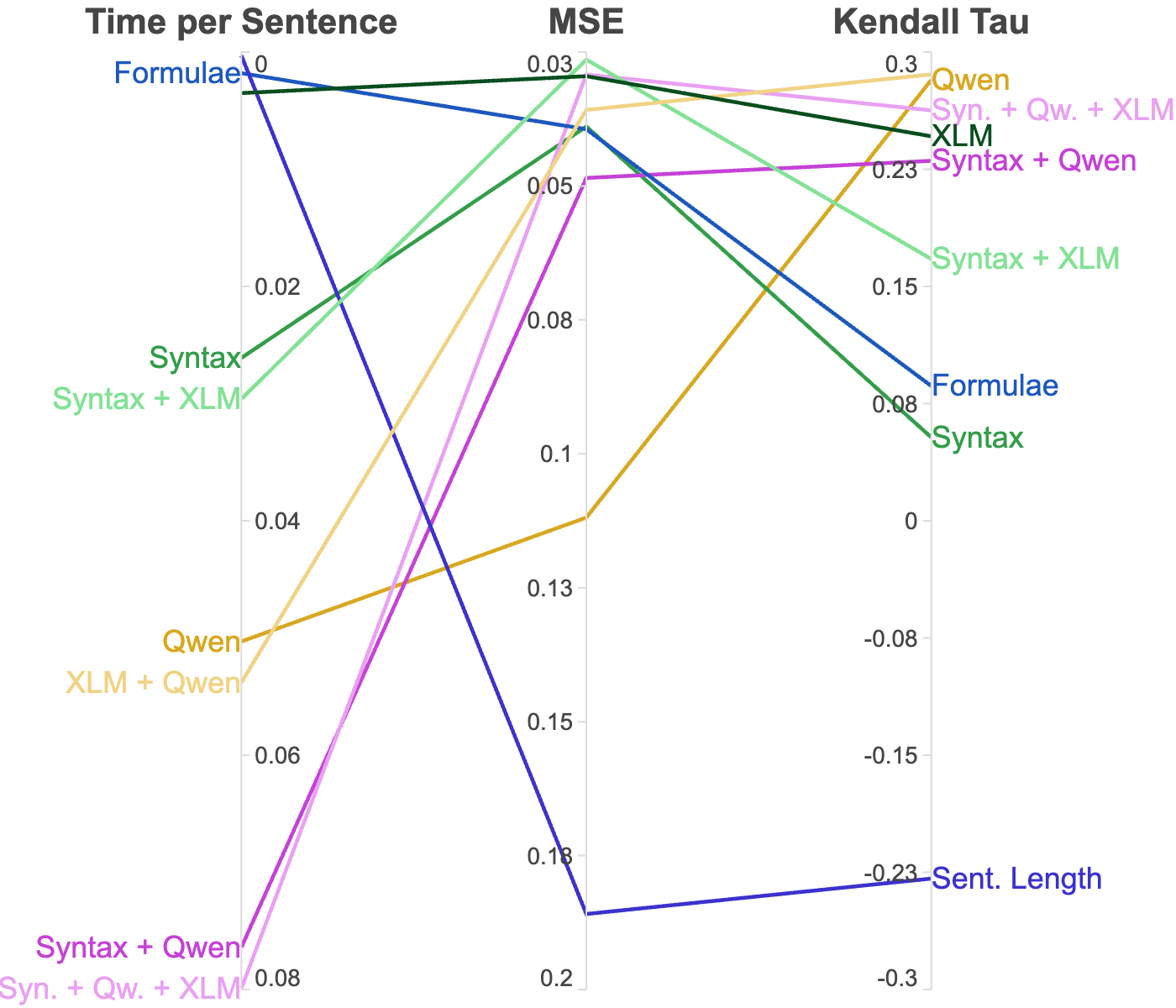}
    \caption{Time-performance trade-offs. All axes rank models top-down from best to worst.}
    \label{fig:MSE_Time_Graph}
\end{figure}

\subsection{Trading off Errors and Speed}
Some models can rate the readability of a sentence faster while making larger prediction errors than others (figure \ref{fig:MSE_Time_Graph}). Three models appear viable for this tradeoff: The readability formulae baseline is very fast while making small errors. Combining syntax and RoBERTa is slower, but makes even smaller errors. The most viable option is simply using RoBERTa, which is almost as fast as the readability formulae approach and makes almost as little errors as the combination. Any combination with an LLM has higher computational cost, slower speed, and larger score deviations.

\section{Discussion}

We set out to analyze how we can automatically measure the readability of German ESG reports in a way that aligns with non-expert human judgments. In doing so, we gained insight into the following two research questions:

\paragraph{RQ1: How readable are German ESG reports?}

Our data analysis reveals that on average laypersons perceive German ESG reports as easy to read. However, some potentially crucial sentences are unclear and there is considerable variation in judgment between readers. While existing research characterizes ESG reports as generally hard to read \cite[e.g.,][]{su16010260}, our sentence-level approach allows a more nuanced evaluation and, thus, enables more nuanced solutions to the problem.

In contrast to similar research that analyzes specific demographics like second language learners (e.g., \citet{vajjala-meurers-2012-improving}), we have no clearly defined target audience. Our approach allows us to analyze how an average German speaker might perceive the readability of German texts. However, without a clearly defined audience, defining rules for what makes a text readable is difficult. We see this reflected in imperfect agreement among annotators.
Therefore, future research could investigate personalized readability systems \cite{benjamin_reconstructing_2012,bailin_linguistic_2001} while considering that not only the grammatical structure of the target sentence, but also its dependence on local and external context might be important.

\paragraph{RQ2: How to model sentence readability?}

We find that more complex models (more parameters) take longer to rate the readability of a sentence and tend to outperform smaller models in terms of the Kendall $\tau$ score. The LLM outperforms the other individual models in this metric, showing that it is better at delineating easy from hard-to-read sentences. However, as the LLM was not specifically tuned to the task like the other models, it performs the worst according to the MSE.
XLM-RoBERTa has the best tradeoff between low MSE and fast inference.

Intuitively and according to the literature \citep[e.g.,][]{collins-thompson_computational_2014,vajjala-meurers-2012-improving}, word choice and lexical complexity play an important role as well. However, we were not able to replicate this effect in our domain and audience in pilot experiments with word frequency features.

\paragraph{Readability depends on the audience.} A central difficulty with estimating readability is that it depends on genre and domain, as well as the audience. Traditional formulae like Flesch Reading Ease involve coefficients finetuned to the educational domain, which we account for by training a new regressor on our German ESG data.
And while ESG reports are likely authored by trained writers who ensure high readability standards for expert readers, expectations may be different for the average consumer.

\paragraph{Empowering Consumers.}
As shown in figure~\ref{fig:teaser_header}, readability is a crucial building block of true consumer empowerment from ESG reporting, next to other important factors like factuality \citep{climate_fever_2020, florstedt-etal-2025-detecting,luo_etal_2025_exaggeration} and verifiability \citep{prange-etal-2025-overview}.
There are likely inter-correlations between these different aspects of how a company's ESG report is written and that company's actual sustainability strategy \citep{bonn_does_2024}. Although the predictions of even the best models analyzed in this work are far from perfect agreement with the annotators, their scores still provide an indication for the readability as perceived by laypeople and substantiate the complexity of the task.

\section{Conclusion}

In this work, we applied different readability scoring methods to German ESG reports. We evaluated these methods using error and rank correlation metrics, as well as their insight into what makes a sentence hard-to-read (whitebox versus blackbox). Our results show that prompting LLMs has the potential to distinguish clear from hard-to-read sentences. However, a small task-specifically finetuned transformer model predicts human readability with the smallest error. Averaging predictions of multiple models can slightly improve the performance at the cost of slower inference. 

Feature-based models and other explainability methods, which we leave to future work, can identify individual linguistic patterns that impact readability. Thereby, future research could contribute to transparency and consumer empowerment, consumer protection and, through more sustainable consumption patterns, environmental protection and the green transition.

\section{Limitations}
Naturally, any model is an abstraction of reality. Thus, our models are also limited in several ways. Other hyperparameters, LMs, and more complex prompt engineering could lead to different results. Further behavioral and mechanistic explainability methods could allow more thorough investigations of the whitebox (e.g. gradient-based) and even blackbox models (e.g. discretization-based). Finally, analyzing German ESG reports on the document level could be interesting, especially regarding the coherence between adjacent sentences.
Furthermore, we identify the following two major problems:

\paragraph{Difficulties in assessing readability.}

During our experiments, we find that complex context sentences can influence the perception of readability of consecutive sentences. Although the human annotators were tasked to only rate the target sentence, they were able to see the context sentences which might have impacted their ratings. However, our readability models were not able to see the context, leading to an information asymmetry. This poses a general problem to the task of sentence-level readability assessment.

Furthermore, we see a high level of subjectivity in the annotations as seen in the mediocre agreement on the readability ratings.
To limit the influence of outliers, we use the majority vote to aggregate the individual annotations into a single gold truth. However, as the provided examples show, a strong disagreement can influence the majority vote drastically as well. To solve this problem, \citet{benjamin_reconstructing_2012} proposes a personalized readability model trained on the user's browser behavior. However, privacy concerns arise when tracking such personal data. We assume that for only a few users the benefits would outweigh the risks. 

\paragraph{Class imbalance.} Our data shows a strong class imbalance towards easier-to-read sentences. Given that we only analyze the readability as perceived by native speakers, this can be argued with their fundamentally good understanding of German sentences. Furthermore, crowd-workers might fear getting rejected from the task and thus paid less if they admit to not understanding the task, or may simply overestimate themselves. This bias towards very easy sentences is in contrast to existing research that describes ESG reports as ambiguous \cite{bingler_how_2024} and more complex than financial reports \cite{smeuninx_measuring_2020}. This discrepancy might have to do with our focus on sentence-level rather than document-level readability, and may either be a true effect or an artifact of how judgments were collected. We invite future research to replicate and compare different methodologies.

Future work may also address the class imbalance not only at training time but also at test time. In a simple case, for example, performance can be broken down by gold rating, evaluating instances rated as perfectly clear by all annotators separately from all instances that at least one annotator had at least some trouble understanding.

\section{Acknowledgments}
The authors gratefully acknowledge the HPC resources used during early experiments that were provided by the Erlangen National High Performance Computing Center (NHR@FAU) of the Friedrich-Alexander-Universität Erlangen-Nürnberg (FAU) under the BayernKI project v110ee. BayernKI funding is provided by Bavarian state authorities. 
The majority of the work for this paper was done at the chair of Computational Linguistics of Prof. Dr. Annemarie Friedrich at the University of Augsburg and we are very grateful for their support.
Further, we thank the anonymous reviewers for their constructive feedback.
We also thank Nina Prange for her input on the public communication aspect of consumer empowerment.

\section{Bibliographical References}\label{sec:reference}
\bibliographystyle{lrec2026-natbib}
\bibliography{custom}

\section{Language Resource References}
\label{lr:ref}
\bibliographystylelanguageresource{lrec2026-natbib}
\bibliographylanguageresource{languageresource}

\input{latex/appendix}

\end{document}

%% file: tables/data_table_complete.tex
\begin{table}[t]
    \setlength{\tabcolsep}{1.2mm}
    \centering\small
    \begin{tabular}{l rrr}
    \toprule
         & \multicolumn{1}{c}{\textbf{Train}} & \multicolumn{1}{c}{\textbf{Dev}} & \multicolumn{1}{c}{\textbf{Eval}} \\\midrule
    \# Sentences & 960\phantom{.000} & 267\phantom{.000} & 407\phantom{.000} \\
    \O\ Words / sentence & 16.92\phantom{0} & 17.50\phantom{0} & 17.32\phantom{0} \\
    \O\ Syllables / word & 2.28\phantom{0} & 2.32\phantom{0} & 2.28\phantom{0} \\
    \midrule
    \multicolumn{3}{l}{\textbf{Inter-Annotator Agreement}} \\
    $\geq$ 3 agree & 86.8\% & 72.3\% & 87.7\% \\
    Mode agreement & 70.3\% & 60.4\% & 70.1\% \\
    \midrule
    \multicolumn{3}{l}{\textbf{Readability Annotations [1.0; 4.0]}} \\
    Avg. mean & 3.515 & 3.200 & 3.526 \\
    Avg. standard deviation & 0.505 & 0.691 & 0.501 \\
    Avg. majority vote & 3.695 & 3.431 & 3.709 \\[3mm]
    \multicolumn{3}{l}{\# Actual majority votes}\\
    \hspace{1em}1.0 & 5\phantom{.000} & 0\phantom{.000} & 0\phantom{.000} \\
    \hspace{1em}1.5 & 1\phantom{.000} & 0\phantom{.000} & 0\phantom{.000} \\
    \hspace{1em}2.0 & 21\phantom{.000} & 11\phantom{.000} & 7\phantom{.000} \\
    \hspace{1em}2.5 & 19\phantom{.000} & 14\phantom{.000} & 4\phantom{.000} \\
    \hspace{1em}3.0 & 167\phantom{.000} & 90\phantom{.000} & 81\phantom{.000} \\
    \hspace{1em}3.5 & 76\phantom{.000} & 38\phantom{.000} & 35\phantom{.000} \\
    \hspace{1em}4.0 & 671\phantom{.000} & 114\phantom{.000} & 280\phantom{.000} \\
    \bottomrule
    \end{tabular}
    \caption{Dataset statistics.}
    \label{tab:dataset_metrics}
    \label{tab:agreement}
    \label{tab:understandability_annotations}
    \label{tab:understandability_class_imbalances}
\end{table}

%% file: tables/results_reordered.tex

\begin{table*}[ht]
    \setlength{\tabcolsep}{1.6mm}
    \centering
    \small

    \sisetup{detect-weight=true,detect-inline-weight=math, table-alignment-mode = format}

    \def\Uline#1{#1\llap{\uline{\phantom{#1}}}} 

    \begin{tabular}{
        l 
        l  
        S[retain-explicit-plus, retain-negative-zero, table-format = 1.4] 
        S[retain-explicit-plus, retain-negative-zero, table-format = 1.4] 
        S[retain-explicit-plus, retain-negative-zero, table-format = 1.4] 
        S[retain-explicit-plus, retain-negative-zero, table-format = 1.4s]
        r
    }

    \toprule
     
     \makecell[c]{\textbf{Type}} & \makecell[c]{\textbf{Model}} & {\textbf{MSE} ($\downarrow$)} & {\textbf{MAE} ($\downarrow$)} & {\makecell[c]{\textbf{Kendall $\tau$} \\ ($\uparrow$)}} & {\makecell{\textbf{\O\ Time per} \\ \textbf{Sentence ($\downarrow$)}}} & \makecell[c]{\textbf{\#\  Params}} \\

     \midrule
     
     \multirow[c]{3}{*}{$\square$ Whitebox} & Sentence length baseline & 0.1859 & 0.4017 & -0.2290 & 0.0003s & 1\\


     & Readability formulae baseline & 0.0394& 0.1588 & \uline{0.0863} & \bfseries\uline{0.0018s} & 5 \\


    & Syntactic features (ours) & \underline{0.0389} & \uline{0.1502} & 0.0534 & 0.0261s & $\sim$0.7M \\

     \midrule

     \multirow[c]{2}{*}{$\blacksquare$ XLM-RoBERTa} & base & \uline{0.0295} & \bfseries \uline{0.1114} & \uline{0.2461} & \uline{0.0035s} & $\sim$278M\\

     & large & 0.1325 & 0.3373  & -0.2198 & NA & $\sim$550M \\

     \midrule

     \multirow[c]{3}{*}{$\blacksquare$ LLMs}


      & \makecell[l]{Qwen 3 4B Instruct 2507} & \uline{0.1119} & \uline{0.2469} &  \uline{0.2822} & \uline{0.0503s} & $\sim$4,000M \\

      & \makecell[l]{Gemma 3 4B it} & 0.2396 & 0.4402 & 0.0448 & 0.3110s & $\sim$4,000M\\

      & \makecell[l]{Llama 3 8B instruct} & 0.2347 & 0.4230 & -0.1906 & 0.7100s & $\sim$8,000M \\

     \midrule

    \multirow[c]{4}{*}{Combinations} & Syntax + XLM-base & \bfseries \uline{0.0264} &	\uline{0.1141} & 0.1676 & \uline{0.0296s} & $\sim$279M \\
    & Syntax + Qwen & 0.0485 & 0.1777 & 0.2304 & 0.0764s & $\sim$4,001M \\
    & XLM-base + Qwen & 0.0358 & 0.1432 & \bfseries \uline{0.2857} & 0.0538s & $\sim$4,278M  \\
    & Syntax + XLM-base + Qwen & 0.0292 & 0.1355 & 0.2627 & 0.0799s & $\sim$4,279M \\

    \bottomrule

    \end{tabular}

    \caption{Results of the experiments. $\downarrow$ indicates that a lower value is better and $\uparrow$ indicates that a larger value is better. The best value per metric is bold and the best per model type is underlined. For details on the experimental setup see appendix \ref{appendix:experimental-setup}.}
    \label{tab:results}
\end{table*}

%% file: tables/average_predictions.tex
\begin{table}
    \setlength{\tabcolsep}{1.5mm}
    \centering
    \small
    
    \begin{tabular}{l  c c c}
        \toprule
        & \textbf{Train} & \textbf{Dev} & \textbf{Eval} \\
         \midrule

        \makecell[l]{Sentence length \\ baseline} & 2.523 & 2.535 & 2.531 \\
        \makecell[l]{Readability formulae \\ baseline} & 3.527 & 3.551 & 3.528 \\
        Syntactic features & 3.608 & 3.555 & 3.596 \\
        XLM-RoBERTa base & 3.790 & 3.933 & 3.920  \\
        Qwen 3 & 3.077 & 3.075 & 3.037 \\

        \midrule

        Human Annotation & 3.695 & 3.431 & 3.709 \\

        \bottomrule
    \end{tabular}
    
    \caption{Average model predictions.}
    \label{tab:average_predictions}
\end{table}

%% file: tables/ARA-Syntax-ablation.tex
\begin{table}[]
    \setlength{\tabcolsep}{1.5mm}
    \centering
    \small

    \sisetup{detect-weight=true,detect-inline-weight=math, table-alignment-mode = format}

    \begin{tabular}{
        l |
        S[retain-explicit-plus, retain-negative-zero, table-format = +1.4]
        S[retain-explicit-plus, retain-negative-zero, table-format = +1.4]
        S[retain-explicit-plus, retain-negative-zero, table-format = +1.4]
    }

    \toprule       

     {\makecell{\textbf{Ablated Feature}}} & {\textbf{MSE}} & {\textbf{MAE}} & {\textbf{Kendall $\tau$}} \\

     \midrule

    Sentence Depth & +0.0043 & +0.0282 & -0.0904  \\

	Dependency Dist. & -0.0046 & -0.0104 & -0.0286\\

    Sentence Root & -0.0045 & -0.0321 & -0.0123 \\

	Is Passive & +0.0116 & +0.0329 &  -0.0690 \\

    Has Subordination & +0.0036 & +0.0039 &  -0.0508  \\

	Bigrams & -0.0031 & -0.0033 &  -0.0467\\

	Trigrams & \bfseries +0.0227 & \bfseries +0.0617 &  \bfseries -0.1103  \\

	\midrule

    All Features & 0.0369 & 0.1502 &  0.1203 \\

    \bottomrule
    
    \end{tabular}

    \caption[Ablation of the Syntax-based ARA model on the evaluation data split.]{Ablation of the Syntax-based ARA model on the evaluation data split. We report the differences in metrics to the complete model (last row).}
    \label{tab:results_ARA-Syntax-ablation}
\end{table}

%% file: latex/appendix.tex
\appendix

\section{Systematic Literature Search}
\label{appendix:literature_search}
To find existing research on the readability of German ESG reports, we queried several research databases: We searched in the ACL Anthology\footnote{Available at \url{https://aclanthology.org/}, last accessed Sep 24, 2025.}, which focuses on NLP research. We queried the DBLP\footnote{Available at \url{https://dblp.uni-trier.de/}, last accessed Sep 24, 2025.}, a German computer science bibliography, to include computer science research in general. We searched the EBSCOhost database\footnote{Available at \url{https://research.ebsco.com/}, last accessed Sep 24, 2025.} and Scopus\footnote{Available at \url{https://scopus.com/}, last accessed on 24.9.2025.} to include research from the area of business informatics. Finally, we searched the Web of Science\footnote{Available at \url{https://webofscience.com/}, last accessed Sep 24, 2025.} as it includes research from various disciplines. All databases were queried using with the following search string:

\texttt{(readability OR understandability OR ((text OR sentence) AND (complexity or \\ difficulty))) AND ("Environment* Social* Governance" OR ESG OR "Corporate Social \\ 
Responsibility" OR CSR OR "sustainability report" OR "company climate report") AND German}

This search string thereby combines several terms that describe readability with terms for German ESG reports. We searched the title, keywords, and abstract fields on the 24th of September in 2025.

\section{Implementation Details for the Syntax Model}
\label{appendix:syntax-implementation}

We load the \textit{de\_dep\_news\_trf} POS-tagging model for the German language by the python library \texttt{spacy}. It allows extracting the POS-tags of each word, identifying the root of a sentence and its POS-tag, and extracting the depth of the dependency tree. The mean dependency distance is calculated by an extension to the \texttt{spacy} library called \texttt{textdescriptives}. If \texttt{spacy} detects a passivized subject or the sentence includes a participle with a form of \enquote{werden} as its head, it is considered passive. Finally, a sentence has a subordinate clause if \texttt{spacy} finds a subordinate conjunction.

The features are aggregated in a neural network. The first layer is split into two parts. In the first part, the $n$-grams vector is compressed to 500 neurons to reduce its impact on the model and the remaining features are expanded to 25 neurons for the second part of the first layer. Then, we concatenate the two parts and pass them to the second layer consisting of 256 neurons. The third layer compresses the model down to 128 neurons before the model outputs the regression value in the single output neuron. After each layer, except for the output layer, we add the ReLU activation function \cite{nair_rectified_2010} and 10\% dropout \cite{srivastava_dropout_2014}. The model is trained using the AdamW optimizer \cite{loshchilov_decoupled_2017,kingma_adam_2015}. The following hyperparameters were identified using grid search: batch size: 20; training epochs: 40; learning rate: 0.01; early stopping patience: 15.

\section{Implementation Details for the Readability Formulae Baseline}
\label{appendix:rf-model}

Based on their historical relevance and novelty, we decided to use the following models that are applicable to German sentences:

\paragraph{Flesch-Reading-Ease Test.}
Since its introduction by \citet{flesch_new_1948}, the Flesch-Reading-Ease test has often been used to rate the readability of English sentences \cite{kauchak_measuring_2017}. It calculates a readability score based on the number of words per sentence and number of syllables per word. \citet{amstad_wie_1978} recalculated its factors to fit the formula to German sentences:

\begin{equation}
    \small
    \text{\parbox[c]{1.1cm}{Flesch Reading Ease}} = 180 - \left(\frac{\text{\#\ Words}}{\text{\#\ Sentences}}\right) - 58.5 \cdot \left(\frac{\text{\#\ Syllables}}{\text{\#\ Words}}\right)
\end{equation}

\paragraph{Hohenheim Complexity Index.}

The HKPS \citep{kercher_verstehen_2013} is based on articles on politics from the German newspaper \textit{BILD} and on dissertations on politics from PhD students. Shallow sentence and word features are weight against each other based on their importance in the two text groups. If a text is more similar to a BILD article it is easier-to-read for laypersons, whereas dissertations are harder-to-read for laypersons. 

\paragraph{Polysyllabic Proportion.}

The idea that sentences containing many long words tend to be more complex has been often applied in research. This is, for example, one of the core ideas shared by the SMOG index \cite{mclaughlin_smog_1969} and the Gunning Fox Index \cite{gunning_technique_1952}. However, both are neither designed nor adapted to German texts. Therefore, we use the simple polysyllabic proportion as a feature for our readability formulae-based model and follow \citeauthor{mclaughlin_smog_1969}'s definition of polysyllabic words as words with at least three syllables.

\begin{equation}
\small
\text{Polysyllabic Proportion} = \frac{\text{\#\ Polysyllabic Words}}{\text{\#\ Words}}
\end{equation}

\paragraph{Vienna Educational Text Formula.}

The Vienna formula was specifically designed for German scientific texts by \citet{bamberger_lesen-verstehen-lernen-schreiben_1984}. It considers the proportion of polysyllabic words, the length of sentences, long words, and the proportion of monosyllabic words. The authors supply three versions of the WSTF (Wiener Sachtext Formel in German). We use the first one as it is the most accurate one, according to the authors.

\begin{equation}
\small
\begin{aligned}
\text{WSTF} = 0.1935 \cdot \text{MS} \\ + 0.1672 \cdot \text{Average Words per Sentence}\\
+ 0.1297 \cdot \text{IW} - 0.0327 \cdot \text{ES} - 0.875,
\end{aligned}
\end{equation}

where MS is the percentage of polysyllabic words; IW is the percentage of words with more than six characters; and ES is the percentage of monosyllabic words. The scale represents the expected year of full-time education that is required to understand the text and ranges from 4 (easy) to 15 (very difficult).

\paragraph{Swedish Readability Index.}

Finally, the Läsbarhetsindex (LIX) was designed for the Swedish language by \citet{bjornsson_lasbarhet_1968} and has already been successfully applied to other Germanic languages \cite[e.g.,][]{wold_estimating_2024}.

\begin{equation}
\small
\text{LIX} = \frac{\text{Number of words}}{\text{Number of sentences}} \cdot \frac{\text{Number of long words} \cdot 100}{\text{Number of words}}
\end{equation}

Long words are defined as words with more than six characters. \citet{bjornsson_lasbarhet_1968} also provides a table that assigns readability classes for the score, but we use the raw LIX score to retain all information.

\paragraph{Model-specific implementation details.}
The Flesch Reading Ease and the first Vienna Eucational Text Formula are calculated using the python library \texttt{textstat}. The formulae for the polysyllabic proportion, the LIX score, and the HKPS are implemented by the authors of this work. The XGBoost aggregation method is trained with the MSE objective and the following parameters: number of boosted trees: 100; learning rate: 0.1; maximum tree depth: 5. These parameters were identified using a simple grid search,

\section{Implementation Details for the Experiments with XLM-RoBERTa}
\label{appendix:RoBERTa-implementation}

The RoBERTa checkpoints are loaded using the library \texttt{transformers}. The following hyperparameters were identified using grid search: batch size: 20; training epochs: 30; learning rate: 0.0001; weight decay: 0.001; gradient disabled for the first five layers.

\section{LLM Prompting}
\label{appendix:llm-prompt}

The LLMs are loaded using the \texttt{transformers} library with disabled sampling for reproducibility. They are prompted to rate the readability of sentences as follows: First, a system prompt outlines the general task and indicates to the model that it is supposed to rate the readability of German sentences (see figure \ref{fig:llm-prompt}). Second, a user prompt outlines the structure in which the model is supposed to output. This includes describing the rating scale from 1 to 4 and telling to model to only output a single digit. Furthermore, the user prompt includes a single shot based on the training split of our dataset including the majority vote of the human annotators delineated by placeholder tokens. The user prompt ends with the sentence that is supposed to be rated. Third, the LLM's output begins with a placeholder token for the score, followed by the score generated by the model.

\input{figures/ARA-LLM-oneshot}

\section{Experimental Setup}
\label{appendix:experimental-setup}
Unless otherwise specified, we always use the default hyperparameters.

\paragraph{Hardware.}
The results to the experiments listed in this work were all created in a local workstation with an NVIDIA RTX 5080 with 16GB of VRAM, paired with an AMD 9800x3D CPU. 

\paragraph{Software.}
All experiments were carried out using Python 3.12 with separate Conda environments for each model. CUDA version 12.8 was used. The experiments were run on Ubuntu 24.04 LTS within Windows Subsystem for Linux 2 on Microsoft Windows 11. We always set a seed for reproducibility.

%% file: figures/ARA-LLM-oneshot.tex
\begin{figure}[H]
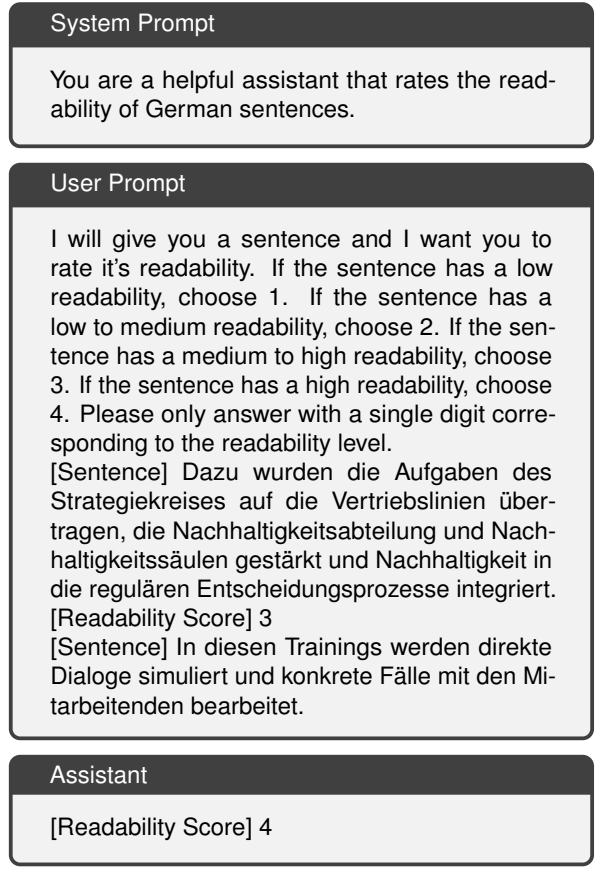

    \small
        
    \begin{tcolorbox}[title=System Prompt]
             
        You are a helpful assistant that rates the readability of German sentences.
        
    \end{tcolorbox}
    
    \begin{tcolorbox}[title=User Prompt]
        I will give you a sentence and I want you to rate it's readability. If the sentence has a low readability, choose 1. If the sentence has a low to medium readability, choose 2. If the sentence has a medium to high readability, choose 3. If the sentence has a high readability, choose 4. Please only answer with a single digit corresponding to the readability level. 
        
        [Sentence] Dazu wurden die Aufgaben des Strategiekreises auf die Vertriebslinien übertragen, die Nachhaltigkeitsabteilung und Nachhaltigkeitssäulen gestärkt und Nachhaltigkeit in die regulären Entscheidungsprozesse integriert. [Readability Score] 3 
        
        [Sentence] In diesen Trainings werden direkte Dialoge simuliert und konkrete Fälle mit den Mitarbeitenden bearbeitet.
    \end{tcolorbox}
    
    \begin{tcolorbox}[title=Assistant]
        [Readability Score] 4
    
    \end{tcolorbox}

    \caption{Prompt for the LLM-based ARA model using single-shot prompting.}
    \label{fig:llm-prompt}
\end{figure}

%% file: main.bbl
\begin{thebibliography}{1}
\expandafter\ifx\csname natexlab\endcsname\relax\def\natexlab#1{#1}\fi

\bibitem[{Prange et~al.(2025)Prange, Jakob, G{\"o}ttfert, Huber, Wenzel~Neves, and Friedrich}]{sustaineval-2025-data}
Prange, Jakob and Jakob, Charlott and G{\"o}ttfert, Patrick and Huber, Raphael and Wenzel Neves, Pia and Friedrich, Annemarie. 2025.
\newblock \href {https://github.com/SustainEval/sustaineval2025_data/} {\emph{{S}ustain{E}val 2025 Data}}.
\newblock Available on GitHub.

\end{thebibliography}
